\documentclass[kdmile, a4paper]{kdmile}

\usepackage{graphicx}
\usepackage[T1]{fontenc}
\usepackage{color}
\usepackage{amsmath, amssymb}
\usepackage{float}
\usepackage{subfigure}
\usepackage{bm}
\usepackage{url}

\newdef{definition}[theorem]{Definition}
\newdef{remark}[theorem]{Remark}

\title{Characterizing instance hardness in classification and regression problems}

\author{
    Gustavo P. Torquette\inst{1}, 
    Victor S. Nunes\inst{2}, 
    Pedro Y. A. Paiva\inst{2},
    Lourenço B. Cunha Neto\inst{2},
    Ana C. Lorena%\footnote{Thanks for FAPESP grant 2021/06870-3}
    \inst{2}
}

\institute{
    Universidade Federal de São Paulo (UNIFESP), São José dos Campos, São Paulo, Brazil \\ \email{gustavo.torquette@unifesp.br}
    \and 
    Instituto Tecnológico da Aeronáutica (ITA), São José dos Campos, São Paulo, Brazil  \\ \email{victor.nunes@ga.ita.br; \{paiva, aclorena\}@ita.br}
}

\begin{abstract}
Some recent pieces of work in the Machine Learning (ML) literature have demonstrated the usefulness of assessing which observations are hardest to have their label predicted accurately. By identifying such instances, one may inspect whether they have any quality issues that should be addressed. Learning strategies based on the difficulty level of the observations can also be devised. This paper presents a set of meta-features that aim at characterizing which instances of a dataset are hardest to have their label predicted accurately and why they are so, aka instance hardness measures. Both classification and regression problems are considered. Synthetic datasets with different levels of complexity are built and analyzed. A Python package containing all implementations is also provided.
\end{abstract}


\graphicspath{ {./images/regression}{./images/classification} }

\begin{document}

% This is optional:
\begin{bottomstuff}
% similar to \thanks
% for authors' addresses; research/grant statements
\end{bottomstuff}

\maketitle

\section{Introduction}

A recent movement in the Machine Learning (ML) community named \textit{data-centric} Artificial Intelligence (AI) claims one needs to focus more efforts on data understanding and data quality improvement than on the development of more complex ML models %\cite{miranda2021towards,
\cite{schweighofer2022data}. Meanwhile, the Metalearning (MtL) literature is plenty of measures, also named meta-features, able to characterize different properties of a dataset \cite{rivolli2022meta}. Among the existent meta-features, those able to estimate the difficulty level of a dataset, aka \textit{data complexity measures}, have proven to provide valuable information on %the geometric and topological structure of a dataset 
the underlying challenges posed by different classification \cite{lorena2019complex} and regression \cite{lorena2018data} datasets. 

For classification problems, Smith et al. [2014] have also introduced a set of measures able to characterize the difficulty level of each individual observation (instance) of a dataset. Some of the data complexity measures can also be decomposed for an instance-level analysis %as HM 
\cite{arruda2020measuring}. They are named \textit{instance hardness measures} (HM) and try to pinpoint those observations of a dataset that any ML technique will probably struggle to classify correctly and possible reasons why they are hard to classify, giving insights for supporting a data-centric analysis. For regression problems, on the other hand, there are no similar previous studies on measuring the hardness level of the individual observations of a dataset based on HM. This paper tries to bridge this gap by proposing a set of HM for regression problems. For this, we also propose a definition of how difficult it can be to predict the label of an instance in regression problems. Next, we perform a systematic analysis of the HM of classification and regression problems by %defining, 
generating and analyzing synthetic datasets of different complexity levels. Finally, we provide a Python package containing implementations of all HM presented.

This paper is structured as follows: Section \ref{sec:definitions} presents the background on measuring instance hardness. Section \ref{sec:proposal} presents the HM of classification and regression problems. Section \ref{sec:experiments} presents an experimental evaluation of the HM. Section \ref{sec:conclusion} concludes this paper.

\section{Instance Hardness Definition} \label{sec:definitions}

While many studies in the field of MtL focus mainly on dataset level analysis, few approach instance level investigation in a comprehensive way, where instead of comparing the difficulty of different datasets, one is interested in investigating how hard instances belonging to the same dataset are to classify. %Topics such as outlier detection and class imbalance, for example, are aspects that may explain why some instances are harder than others, but they do not cover all the possible cases. 
A more systematic study is presented in \cite{smith2014instance}, where \textit{instance hardness} (IH), roughly speaking, is defined as the likelihood of an instance being misclassified despite the ML technique considered. Given a pool of classifiers $\mathcal L$ with different biases, they define the IH of an instance as:

\begin{equation}\label{eq:IH}
    IH_{\mathcal{L}}\big( \mathbf x_i, y_i \big) = 1 - \frac{1}{|\mathcal{L}|} \sum_{j=1}^{|\mathcal{L}|} p\big(y_i | h_j(\mathbf x_i)\big),
\end{equation}
where $p\big(y_i | h_j(\mathbf x_i)\big)$ is the probability assigned to $\mathbf x_i$ to its original label $y_i$ in the dataset by a learning model $h_j$ in the pool. The idea %in Equation \ref{eq:IH} 
is that instances that are frequently misclassified by a pool of diverse learning algorithms can be considered hard. On the other hand, easy instances are likely to be correctly classified by any of the considered algorithms.

%\subsection{Instance Hardness for regression}
While the definition of instance hardness for classification problems arises naturally, we cannot say the same for regression problems. Discrete target values induce a discrete probability space, if we set the sample space $\Omega = Y$ for a discrete output space. For classification problems the categorical outputs make it possible to readily measure the probabilities assigned to each of the classes. In regression problems, however, the set $Y$ is continuous, implying that adaptations are needed. Disregarding domain knowledge, how can one tell if a prediction made by a regression algorithm for a given instance is good or bad? This question is quite tricky, but answering it would give hints on how to define IH for regression. One possible answer is to define a relative measure of IH, in which case some instances in a dataset would be considered easy and some hard. That is as far as we can get without any prior knowledge, but may be sufficient for datasets displaying non-zero variance on the response variable, $\mathrm{var}(\bm y) > 0$.

It is a fact that the closer the predicted value is to the actual label of an instance, the more accurate the regressor response is. Therefore, it is more intuitive to define our probability space over distances $z = \mathrm{d}(y, \hat{y}_h)$, where $y$ is the label of an instance and $\hat y_h$ is the prediction obtained by a model $h$.  We use a kernel $\kappa(z)$ to quantify the similarity. For instance, consider an exponential kernel, such that $
%\begin{equation}
    \kappa(z) = \exp\bigg(-\frac{z}{\gamma}\bigg)$. 
%\end{equation}
Plugging it into the cumulative distribution function, %: 
%$$\mathbb{P}(0 < Z \leq z) = F(z) = 1 - \kappa(z),$$
%
%\noindent then,
%\begin{align*}
 %   f(z) &= \frac{\mathrm{d}}{\mathrm{d}z}F(z)\\
  %  &= \frac{\mathrm{d}}{\mathrm{d}z} \Bigg[1 - \exp\bigg(-\frac{z}{\sigma}\bigg)\Bigg]\\
   % &= \frac{1}{\sigma} \exp\bigg(-\frac{z}{\sigma}\bigg).
%\end{align*}
%\noindent If we set $\lambda = \frac{1}{\sigma}$, this function 
we arrive %assumes 
in the form of the exponential distribution. Now, we can define the instance hardness for regression as:

\begin{equation}\label{eq:IHreg}
    IH_{\mathcal{L}}\big( \mathbf x_i, y_i \big) = 1 - \frac{1}{|\mathcal{L}|} \sum_{j=1}^{|\mathcal{L}|} \exp\Bigg(-\frac{\mathrm{d}(y_i, \hat{y}_{ji})}{\gamma}\Bigg),
\end{equation}
where $\hat y_{ji}$ represents a regressor $j$ in the pool $\mathcal L$ makes for instance $\mathbf x_i$. 
A natural issue that arises is how to choose a proper value for $\gamma$. We can see it as a normalizing constant. For instance, we can set $\gamma$ as the power of the signal $\mathbf y$, that is, $\gamma = \frac{1}{N} \sum_i y_i^2$. In that case, depending on the chosen distance metric $\mathrm{d}(\cdot, \cdot)$, $\frac{\mathrm{d}(\cdot, \cdot)}{\gamma}$ is equivalent to some normalized error metric, such as normalized squared error when the Euclidean distance is used.

\section{Instance Hardness Measures}\label{sec:proposal}

Apart from the definition of IH, Smith et al. %\cite{smith2014instance}
[2014] also propose a set of \textit{hardness measures} (HM) for classification problems, which aim to explain \textit{why} some instances are harder to have their label predicted correctly than others in a dataset. These measures are presented next, along with novel HM devoted to regression problems. All of them are computed for the instances of a dataset $D$ with $n$ data instances $\mathbf x_i$ assuming labels $y_i$ in a set $Y$, which is qualitative for classification problems and quantitative for regression problems. The definitions of all measures are %computed using the entire dataset. In this paper they are all 
standardized in this paper so that larger values are observed for more difficult instances. 

\subsection{HM for Classification}\label{sec:clas}

The HM for classification used here were defined in \cite{smith2014instance,arruda2020measuring}, except from the last one. There are also other HM in these works, but they have similar definitions and are redundant to at least one of the presented measures and are omitted here.  

\begin{description}
    
    \item[\textbf{k-Disagreeing Neighbors}] $kDN(\mathbf x_i)$: %\cite{smith2014instance}: 
    outputs the percentage of the $k$ nearest neighbors of $\mathbf x_i$ in $D$ which do not share its label, 
   with $k$ set to 5. % \cite{smith2014instance}. 
   Instances %The higher the value of $kDN(\mathbf x_i)$, the harder $\mathbf x_i$'s classification tends to be, since it is 
   surrounded by examples from a different class are harder to classify. 
    
    \item[\textbf{Disjunct Class Percentage}] $DCP(\mathbf x_i)$:  %\cite{smith2014instance}: 
    this HM  builds a decision tree using $D$ and considers the percentage of instances in the disjunct of $\mathbf x_i$ which share the same label as $\mathbf x_i$. 
    Easier instances will have a larger percentage of examples sharing the same label as them in their disjunct, so we output the complement of this percentage. 
    
    \item[\textbf{Tree Depth}] $TD(\mathbf x_i)$: % \cite{smith2014instance}: 
    depth of the leaf node that classifies $\mathbf x_i$ in an unpruned decision tree, normalized by the maximum depth of the tree built from $D$. 
    Harder to classify instances tend to be placed at deeper levels of the tree and present higher $TD$ values. 
    
    \item[\textbf{Class Likelihood Difference}] $CLD(\mathbf x_i)$: % \cite{smith2014instance}: 
    takes the difference between the likelihood that $\mathbf x_i$ belongs to its class $y_i$ and the maximum likelihood it has to any other class. 
    The difference in the class likelihood is larger for easier instances, because the confidence it belongs to its class is larger than that of any other class. We take the complement of the difference.
    
    \item[\textbf{Class Balance}] $CB(\mathbf x_i)$:  %\cite{smith2014instance}: 
    measures the skewness of $\mathbf x_i$'s class by taking the proportion of instances that share the label of $\mathbf x_i$ in  $D$. 
    We take a complement of the measure so that $CB$ will be minimum for all instances if the problem is balanced, which is simpler concerning the class balance aspect.
    
    \item[\textbf{Fraction of features in overlapping areas}] $F1(\mathbf x_i)$: % \cite{arruda2020measuring}: 
    this measure takes the percentage of features of the instance $\mathbf x_i$ whose values lie in an overlapping region of the classes. 
    One may regard a feature as having overlap if it is not possible to separate the classes using a threshold on that feature's values.  %$F1$ gives the percentage of features for which an example is in an overlapping region. 
    Larger values of $F1$ are obtained for data instances which lie in overlapping regions for most of the features, implying they are harder to classify according to the $F1$ interpretation. 
    
    \item[\textbf{Fraction of nearby instances of different classes}] $N1(\mathbf x_i)$: % \cite{arruda2020measuring}: 
    first a minimum spanning tree MST is built, where each instance of the dataset $D$ corresponds to one vertex and nearby instances are connected according to their distances %in the input space 
    in order to obtain a tree of minimal cost concerning the sum of the edges' weights. %so that each instance corresponds to one vertex and is connected to nearby instances. 
    $N1$ gives the percentage of instances of different classes $\mathbf x_i$ is connected to. 
    Larger values indicate that $\mathbf x_i$ is close to examples of different classes, %either because it is borderline or noisy, 
    making it hard to classify. 
    
    \item[\textbf{Ratio of the intra-class and extra-class distances}] $N2(\mathbf x_i)$: considers the complement of the ratio of the distance of $\mathbf x_i$ to the nearest example from its class to the distance it has to the nearest instance from a different class (nearest enemy). 
    Larger values of $N2$ indicate that $\mathbf x_i$ is closer to an example from another class than to an example from its own class and is harder to classify. 
    
    \item[\textbf{Local Set Cardinality}] $LSC(\mathbf x_i)$:  %\cite{arruda2020measuring}: 
    the Local-Set (LS) of an instance $\mathbf x_i$ is the set of points from $D$ whose distances to $\mathbf x_i$ are smaller than the distance between $\mathbf x_i$ and $\mathbf x_i$'s nearest enemy \cite{Leyva:set}. 
    $LSC$ outputs the relative cardinality of such set, so that 
    larger local sets are obtained for easier examples, which are in dense regions surrounded by instances from their own classes. %In Equation \eqref{eq:LSC} 
    For standardization, we output a complement of the relative local set cardinality. 
    
    \item[\textbf{Local Set Radius}] $LSR(\mathbf x_i)$: % \cite{arruda2020measuring}: 
    takes the normalized radius of $\mathbf x_i$'s local set. 
    Larger radiuses are expected for easier instances, which are surrounded by many instances sharing its class, so we take the complement of such measure. 
    
    \item[\textbf{Usefulness}] $U(\mathbf x_i)$: % \cite{arruda2020measuring}: 
    fraction of instances having $\mathbf x_i$ in their local sets. 
    If $\mathbf x_i$ is easy to classify, it will be close to many examples from its class and therefore will be more useful. We take the complement of this measure as output. 
    
    \item[\textbf{Density}] $De(\mathbf x_i)$: this measure is proposed here based on a complexity measure originally taken at a dataset-level \cite{lorena2019complex}. First a graph $G=(V,E)$ is built from $D$, connecting instances from the same class for which the distance is inferior to a threshold $\epsilon$, set as 15\% of the smallest distances. % as in \cite{morais2013complex}.  
    The complement of the density of the connections a vertex has in the graph gives its hardness level. 
   If $\mathbf x_i$ is easy to classify, it will be surrounded by close elements from its class and will have a lower $De$ value.% according to Equation \ref{eq:D}.

\end{description}

\subsection{HM for Regression}\label{sec:reg}

For proposing the HM for regression, we took as basis the set of data complexity measures and meta-features for regression problems \cite{lorena2018data,amasyali2009study} and studied how to decompose them at the instance-level. %Part of the measures originally proposed for classification problems presented in the previous subsection can also be adapted for regression problems. 
The measures %$DS$, 
$TD$ and $De$ from the previous section can also be applied to regression problems. While %$DS$ and 
$TD$ will need a regression tree to be induced from $D$ instead of the decision tree, $De$ will take a proximity graph between all instances in $D$, but with no post-processing step for pruning edges.

\begin{description}

\item[\textbf{Collective Feature Efficiency}] $CFE(\mathbf x_i)$: this measure starts by identifying the feature with highest correlation to the output in $D$. All examples with a small residual value ($|\varepsilon_i| \leq 0.1$) after a linear fit between this feature and the target attribute are removed. Then, the most correlated feature to the remaining data points is found and the previous process is repeated until all features have been analyzed or no example remains. For an instance $\mathbf x_i$, we take the round $l_i$ where it is removed from the analysis, normalized by the maximum number of rounds. 
Higher $CFE$ values are obtained for harder instances, which require more features to get a linear fit. 

\item[\textbf{Absolute Error after Linear fit}] $LE(\mathbf x_i)$: first a statistical model of a Multiple Linear Regression is fit to $D$. For each $\mathbf x_i$ a residual or error $\varepsilon_i$ in relation to the actual output $y_i$ can be measured and $LE$ is given as 
$|\varepsilon_i|$. 
Larger values are attained for harder instances.

\item[\textbf{Output Distribution}] $S1(\mathbf x_i)$: As in N1 for classification, first a MST is generated from input data. 
Next S1 monitors whether the instances joined in the MST have similar output values. 
As an instance $\mathbf x_i$ can have multiple neighbors in the MST, we take the average of the differences between their outputs. 
Higher values will be obtained for harder instances, which are connected to examples with dissimilar outputs.

\item[\textbf{Input Distribution}] $S2(\mathbf x_i)$: $S2$ first orders the data points according to their output values $y_i$ and then computes the Euclidean distance between pairs of examples that are neighbors. 
In the ordering, each element will have one or two neighbor examples. For two neighbors, the average should be taken. Otherwise, %if the example is the first or the last in the ordering, 
the unique neighborhood difference is output. 

\item[\textbf{Squared Error of k-nearest neighbor}] $S3(\mathbf x_i)$: calculates the squared error (SE) of a \emph{k-nearest neighbor regressor} (NN), using \emph{leave-one-out}. As in $kDN$, the value of $k$ is set to 5. 

\item[\textbf{Histogram bin}] $HB(\mathbf x_i)$: first a histogram of the normalized values of $\mathbf y$ is taken. Next, we output the complement of the percentage of instances that have their labels placed in the same bin as $\mathbf x_i$. Easy instances will tend to share labels with many other instances in $D$ than hard instances.

\end{description}
%\newpage

\section{Experimental Evaluation and Results}\label{sec:experiments}

In this section we perform some experiments to show how the HM behave for classification and regression datasets of increasing complexities. All measures described previously are implemented in Python and distributed in the PyHard library\footnote{\url{https://pypi.org/project/pyhard/}} \cite{paiva2022relating}.

\subsection{Classification datasets}

Using the mlbench package \cite{leisch2009package} from the R programming language, we generated synthetic datasets where the spread of the data and overlap of the classes continuously increases. The base dataset is the 2D-normals (2-dimensional Gaussian Problem) with two classes and 500 observations, where each class is described by a two-dimensional gaussian with centers equally spaced on a circle around the origin with radius $r = \sqrt{2}$. The standard deviation ($sd$) of the gaussians was increased from 0.1 to 2, at steps of 0.1. The larger the $sd$, the higher the difficulty of the problem. 

\begin{figure}[htp]
    \centering
    \includegraphics[height=.81\textheight]{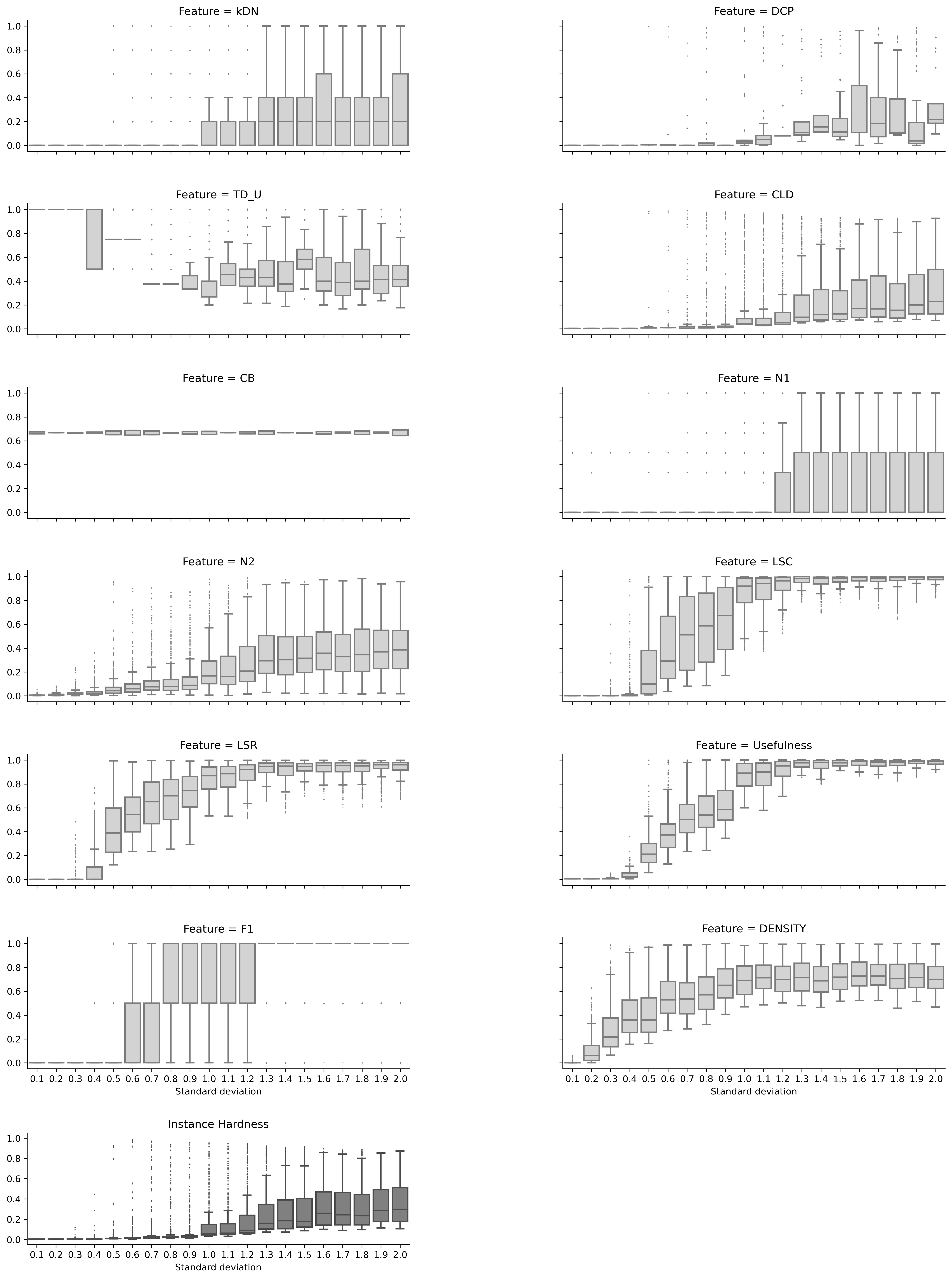}
    \caption{HM for classification and instance hardness values, for all classification datasets generated.}
    \label{fig:hm_im_mix}
\end{figure}

These datasets were input to the PyHard tool to obtain the HM listed in Section \ref{sec:clas}. The HM values obtained for each instance of the datasets were registered and their boxplots are plotted in Figure \ref{fig:hm_im_mix}. Each plot, except for the last, presents the values of one HM ($y$ axis) along the different datasets ($x$ axis), with increasing complexities. We can notice that for most of the HM %some hardness measures like kDN, DS, DCP, CLD, N1, N2, LSC, LSR, Usefulness, F1, Density and Decision Stump presented an 
the values tend to increase when the standard deviation values of the datasets rise, as expected. This behavior is much more evident and smooth for measures $CLD$, $N2$, $LSC$, $LSR$, $U$ and $De$. As the overlap and spread of the classes increases, the instances tend to present a lower likelihood of belonging to their classes, so that higher values for $CLD$ are obtained. This also happens for $N2$, where for datasets with low $sd$ values the classes are compact and the intra class distances are much lower than the inter class distances and this rate inverts for higher $\sigma$ values. $LSC$ and $LSR$ reflect the fact that the classes have increasing overlap for datasets with higher $sd$ values, so the local sets of each instance become smaller. As a reflection of the decrease of the sizes of the local sets, the instances tend to present higher $U$ values for datasets with more overlap too. And $De$ increases as $sd$ increases as a consequence of a higher spread of the data points in the input space, making each class less dense and consequently also decreasing the neighborhood of each instance. 

For $kDN$, $F1$ and $N1$ the results vary less, as these measures present a discrete set of values. $kDN$, for instance, has as possible values for  each instance the values 0, 0.2, 0.4, 0.6, 0.8 and 1. For datasets with few or no overlap ($sd$ from 0.1 to 0.9), the values of $kDN$ are null, as expected, as all instances are surrounded by elements from their own classes. Similarly, $N1$ is also close to 0 for datasets with no or few overlap ($sd$ equal to 0.1 up to 1.1). $F1$ is null for all datasets with low $sd$ values and then increases until becoming dominated by maximum values. Therefore, there is an increase in the overlap of the features as $sd$ increases. $CB$ does not show much variation, because all datasets are built similarly and have similar class distributions. $DCP$ shows a more erratic behavior for datasets with increasing class overlap compared to the previous measures, but the tendency of low values for datasets with no overlap is kept. Finally, some counter-intuitive behaviors are observed for $TD$. When there is no overlap ($sd$ between 0.1 and 0.5), it might be the case one unique split of the dataset and therefore one unique node was enough for classifying most of the dataset correctly. Because of the normalization by the highest depth of the tree, % in Equation \ref{eq:TD}, 
all values become 1 or close to that. But other erratic results for both $DCP$ and $TD$ can also be a consequence on how the decision tree splits the input space using orthogonal cuts on the features values. Many splits may be required even for simple data conformations (for example, for a linearly separable dataset requiring an oblique hyperplane for the separation of the classes).    %The other measures did not presented any relavant trend.

The generated datasets were also fed into seven classification techniques of distinct biases, namely: Support Vector Machine (SVM) with linear Kernel, SVM with RBF Kernel, Random Forest (RF), Gradient Boosting (GB), Bagging, Logistic Regression and Multilayer Perceptron (MLP). All classification techniques were run in a 10-fold cross-validation procedure and had hyperparameters tuned by an inner 3-fold cross-validation on the training folds. Taking these classifiers as the pool $\mathcal L$ in Equation \ref{eq:IH}, the last graph from 
Figure \ref{fig:hm_im_mix} presents the boxplots of the IH values for each of the datasets. As in the case of the HM, we can clearly see a upward trend when increasing the standard deviation from 0.1 to 2. That is, the pool of classifiers tend to present more difficulties in classifying correctly instances of datasets with more spread and overlap. Meta-features such as $CLD$, $N2$, $LSC$, $LSR$, $U$ and $De$  have a high correlation to the instance hardness values as measured by the performance of the pool of classifiers and are good descriptors of the increasing instance hardness observed.% in these datasets.

\subsection{Regression datasets}

The regression datasets were generated as follows \cite{lorena2018data}: each one has 500 observations and one input feature with values 
%Each synthetic dataset has $n$ entries $\mathbf x_i \in \Re^d$, whose $d$ feature values are 
randomly chosen in $[0,1]$ according to an uniform distribution. Next the $y_i \in \Re$ output values are calculated so that the underlying function relating the input to the output is linear. 
Starting from this perfect linear relationship between the input and the output values, we associate a gaussian noise $N(0,\sigma)$ to the labels. The higher the value of $\sigma$, more difficult instances are contained in the dataset. The $\sigma$ values were varied from 0.1 to 1, at steps of 0.1. 

\begin{figure}[htp]
    \centering
    \includegraphics[width=\textwidth]{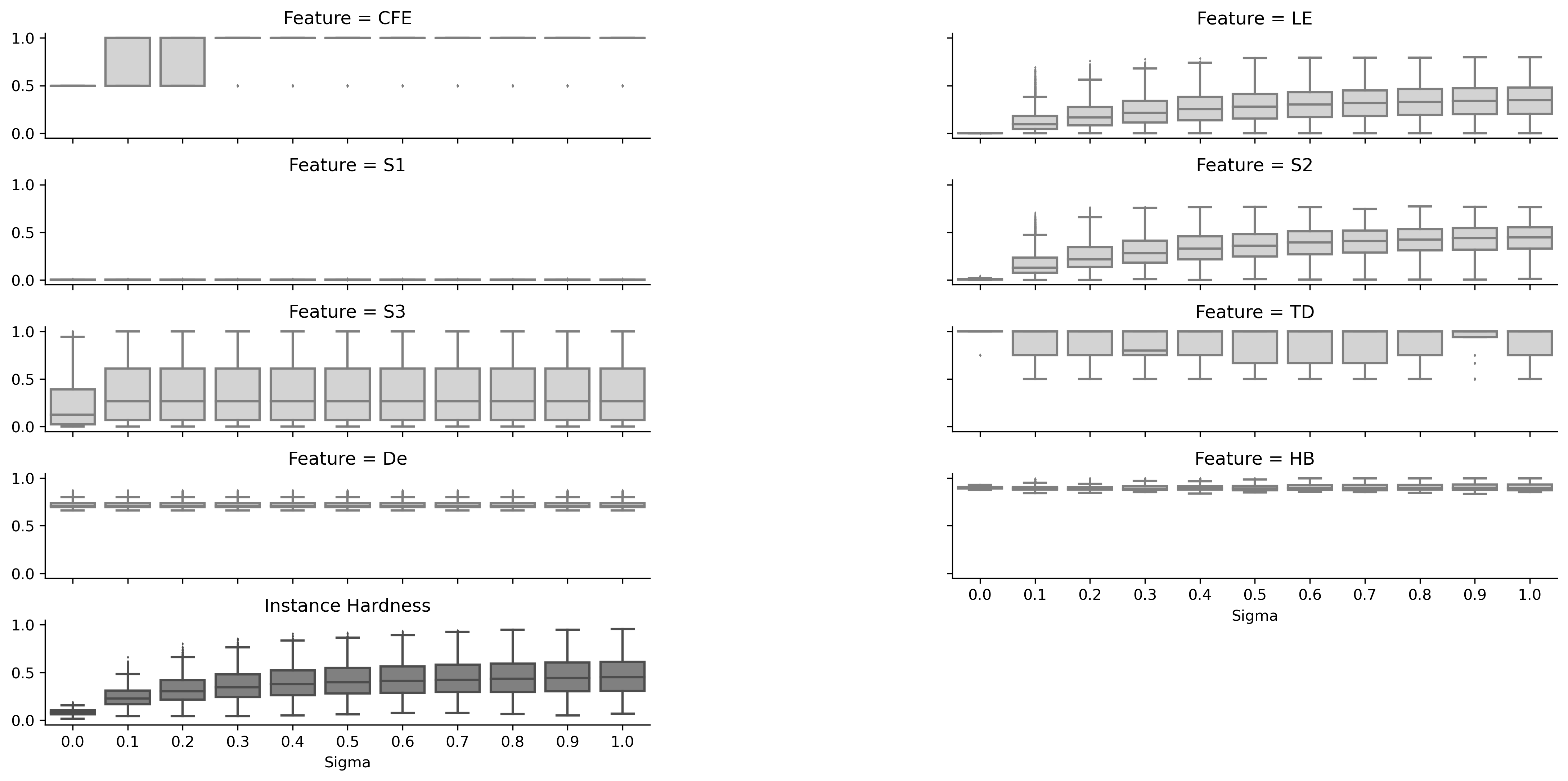}
    \caption{Hardness Measures for regression and Instance Hardness.}
    \label{fig:hm_regression}
\end{figure}

The same procedures adopted for the classification datasets are repeated here. The boxplots of the HM are presented in Figure \ref{fig:hm_regression}, except from the last plot. We can notice that the HM $LE, HB \textnormal{ and } S2$ are able to capture the hardness raising trend as the value of $\sigma$ increases. Since the datasets have primarily a linear structure, which is disturbed by the $\sigma$ values, the increase in the residuals of a linear regressor as monitored by $LE$ was clearly expected. As the $\sigma$ increases, the datasets tend to present a larger spread in the $\mathbf y$ values, justifying why $HB$ is also able to capture these disturbances, although with a lower sensibility. Similarly, neighbor instances in the input space will tend to present large deviation on the labels values for datasets with increasing $\sigma$ values, reflecting in the raise of the $S2$ values accordingly. 

Measure $De$ was not as effective for regression problems compared to the classification counterpart and some alteration on the definition of the proximity graphs might be needed. $S1$ has always shown low values, which may be a consequence of taking the averages of very low differences. % in Equation \ref{eq:S1HD}.
$CFE$ has a tendency similar to that of $F1$ for classification problems. It assumes discrete values and for $\sigma = 0$ all instances correlate perfectly to the output, as expected, as there is a perfect linear relationship between them. But for $\sigma = 0.3$ to 1, for most of the instances the input feature does not correlate to the output and the maximum value is output. $S3$ has a low variation of values and the kNN regressor was not able to capture variations of increasing $\sigma$ values. %$DS$ did not reflect the increasing hardness level of the instances too and all datasets presented similar $DS$ values. 
And $TD$ has presented a very erratic behavior in the regression datasets. This result may be explained by both the normalization factor used for computing this measure and by the bias of the regression trees, which are not adequate for datasets where the linear relationship to the output is oblique. 

The instance hardness boxplots as measured by Equation \ref{eq:IHreg} for the regression datasets is shown in the last plot of Figure \ref{fig:hm_regression}. The pool $\mathcal L$ of regression techniques used in Equation \ref{eq:IHreg} was: AdaBoost, %SVM with linear and RBF kernel, 
$\nu$-SVM, RF, Extremely Randomized Trees, Regression Tree, GN, MLP, Bagging, Bayesian Automatic Relevance Determination, Kernel Ridge Regression, Stochastic Gradient Descent Regression and Passive-Agressive Regression. As we can see in the plot, %figure \ref{fig:ih_regression}, 
the IH values follow an increasing trend as $\sigma$ increases. Therefore, as the $\sigma$ increases, the instances in the corresponding datasets become more difficult to fit, despite the regressor employed. This also confirms that the proposed formulation for IH in Equation \ref{eq:IHreg} captures well the hardness level of the instances. The HM following the trend observed for the IH the most are $LE$ and $S2$.

\section{Conclusions}\label{sec:conclusion}
This paper presents how the hardness level of the individual observations of a dataset in ML can be measured, for both classification and regression problems. First the concept of instance hardness according to the predictive performance of different models is formalized. Next, hardness measures present different perspectives on why an instance is more difficult than another. %For regression problems, this kind of analysis is still absent in the literature.
Experimentally, some of the HM were more effective in reflecting the increase in the difficulty of the instances in a set of synthetic datasets. And measuring the instance hardness of regression problems using a pool of regressors was validated, as the instance hardness values increase for noisy datasets. 
As future work, we must evaluate the measures on other datasets, in particular introducing other sources of complexity than those tested in this paper. For instance, more input features can be added, imposing a higher sparsity in the dataset. The number of classes can also be progressively increased in the case of classification problems. It is also important to validate the use of the measures in real datasets in the future. And some HM should be refined, especially for regression problems. Finally, validating the usage of the HM in applications such as data pre-processing, curriculum learning and active learning are research paths worth future investigations.

\begin{ack}
The authors would like to thank the financial support of FAPESP under grant number 2021/06870-3.
\end{ack}

\bibliographystyle{kdmile}
\bibliography{references}

\end{document}